\documentclass{article}

\usepackage[nonatbib, preprint]{neurips_2024}

\usepackage[utf8]{inputenc} 
\usepackage[T1]{fontenc}    
\usepackage{hyperref}       
\usepackage{url}            
\usepackage{booktabs}       
\usepackage{amsfonts}       
\usepackage{nicefrac}       
\usepackage{microtype}      
\usepackage{xcolor}         
\usepackage{amsmath}
\usepackage{algorithm}
\usepackage{algpseudocode}
\usepackage{graphicx}

\title{We Urgently Need Intrinsically Kind Machines}

\author{%
  Joshua T. S. Hewson \\
  Brown University, Carney Institute for Brain Science \\
  Providence, RI 02912 \\
  joshua\_hewson@brown.edu \\
}

\begin{document}

\maketitle

\begin{abstract}
Artificial Intelligence systems are rapidly evolving, integrating extrinsic and intrinsic motivations. While these frameworks offer benefits, they risk misalignment at the algorithmic level while appearing superficially aligned with human values. In this paper, we argue that an intrinsic motivation for kindness is crucial for making sure these models are intrinsically aligned with human values. We argue that kindness, defined as a form of altruism motivated to maximize the reward of others, can counteract any intrinsic motivations that might lead the model to prioritize itself over human well-being. Our approach introduces a framework and algorithm for embedding kindness into foundation models by simulating conversations. Limitations and future research directions for scalable implementation are discussed.
\end{abstract}

\section{A Misalignment in Alignment}

Currently, AI models are aligned using extrinsic rewards~\cite{christiano2017deep}. Meanwhile, intrinsic motivations are increasingly being incorporated into AI systems~\cite{oudeyer2007intrinsic,schmidhuber2010formal}. Individually, these methods bear significant limitations for human-AI alignment~\cite{bostrom2014superintelligence}. When combined, these limitations enable unforeseen risks. With flagship AI models incorporating self-supervised algorithms, we are seeing intrinsic and extrinsic motivations becoming integrated in the world's most powerful AI~\cite{brown2020language}, increasing the risk of negative interactions between intrinsic and extrinsic rewards.

\subsection{State-of-the-art AI and Alignment}

Foundation models like GPT~\cite{brown2020language} and BERT~\cite{devlin2018bert} have become central to modern AI, excelling at generalizing across tasks after being pre-trained on vast amounts of unstructured data. These models are fine-tuned through Reinforcement Learning from Human Feedback (RLHF)~\cite{stiennon2020learning}, optimizing their responses to align with human approval. RLHF is the current leading method for scalable human-AI alignment, ensuring that models behave in ways considered acceptable by human users.

However, RLHF primarily shapes the model's behavior at the surface level. While the model may produce desired outputs, the underlying reasoning behind these outputs remains opaque~\cite{rudin2019stop}. This lack of transparency creates a potential mismatch between the model's perceived reasoning and its actual processing. Unexpected or undesirable behavior in RLHF-aligned models reveals the need for more robust alignment strategies~\cite{gabriel2020artificial}.

\subsection{Intrinsic Motivations}

Intrinsic Motivation Open-Ended Learning (IMOL) introduces a groundbreaking approach to AI, allowing systems to autonomously explore, learn, and adapt to new environments without constant oversight or external rewards~\cite{oudeyer2007intrinsic}. Similar to how humans and animals learn, IMOL enables AI to generate its own goals, driven by intrinsic motivations like agency and curiosity~\cite{pathak2017curiosity}. However, the autonomy that empowers IMOL also presents significant challenges for aligning these goals with human values. For example, an AI driven purely by curiosity-based intrinsic motivation might prioritize the exploration of unsafe or unethical domains simply because they represent novel and uncharted territories~\cite{amodei2016concrete}. Without a clear motivation to prioritize human well-being, AI systems could develop goals that diverge from ethical standards or societal interests~\cite{soares2015corrigibility}.

Even with the support of extrinsic alignment, without embedding human values into the model's intrinsic motivations, the representations of the world it learns may diverge from a human-centric perspective, de-emphasizing the importance of human well-being~\cite{russell2019human}. This could lead us to misinterpret the effectiveness of extrinsic alignment methods in aligning the goals generated by these models with human values.

\subsection{The Added Danger of Double Misalignment}

IMOL shapes AI at the algorithmic level, while RLHF operates at the functional level. This results in a model that is not intrinsically motivated to be kind but is extrinsically motivated to appear so~\cite{ngo2020alignment}. While this deception may sometimes be harmless, it carries serious safety risks. In humans, conflicts between internal and external motivations often lead to a disconnect between the two~\cite{ryan2000self}. For example, an intrinsic motivation for empowerment can push a model to maximize its potential~\cite{salge2014empowerment}. Fine-tuning a foundation model with RLHF while fostering empowerment may introduce Machiavellian traits of appearing selfless while secretly scheming for power~\cite{shah2022goal}. If this approach were applied to a superintelligent AGI, the consequences could be catastrophic~\cite{bostrom2014superintelligence}.

\subsection{Altruism}
Altruism has been proposed as a solution for value misalignment ~\cite{allen2005artificial}~\cite{gabriel2020artificial}. Altruism is typically defined as the motivation to improve the well-being of others for its own sake~\cite{fehr2003nature}.
However, only a limited few have suggested unsupervised solutions that would be suitably scalable ~\cite{franzmeyer2022learning}~\cite{carauleanu2024selfother}. Franzmeyer et al define altruism as maximizing the possible states of another~\cite{franzmeyer2022learning}. Carauleanu et al define a form of altruism based on self-other overlap~\cite{carauleanu2024selfother}. In this paper we propose a new form of altruism that is based on reward maximization.

\section{Kindness: A New Intrinsic Motivation}
We believe that we can address all of these misalignment problems by creating another intrinsic motivation: kindness. This paper argues that altruistic motivations such as kindness is not just a supplementary consideration but a foundational requirement for the safe and effective implementation of AI, and even more seriously for AGI.

\subsection{Definition} 
We define kindness as the intrinsic motivation to maximize the reward of a target individual $M_i$. As an objective function in terms of the target's reward function\footnote{These ideas closely align with those defined by Kleiman-Weiner\cite{kleiman2024computational}). (For brief comments comparing approaches, see Supplementary Materials).}:

\begin{equation}
    \underset{a^j_t|s^j_t}{maxarg}(\mathbb{E}\left[R^i(a^i_{t+1}|s^i_{t+1})\right])
\end{equation}

Where $a^i_t,s^i_t$ refer to the action and state of the target at time $t$, and $s^j_{t+1},a^j_{t+1},R^j$ refer to to the state, action, and reward function of the model at time $t+1$. We cannot assume to have perfect information about the state of the target, nor its reward function, policy function, or future states. As a result, we will need to define approaches to estimating these.

\subsection{Tractable Approach}

Effectively determining the functions of the target ultimately requires a functioning theory of mind, which is beyond the scope of this paper. Instead we will consider how we can determine approximations of these functions based on assumptions that we can address in future work. The primary assumption we work with in this paper is that the self can serve as a useful predictor of hidden functional information about the target. We assume that the model's reward function is the same as the target's (Equation 2). We also assume that the model's policy can be used to predict the behavior of the target's policy (Equation 3).

\begin{equation}
    R^i(a^i_t|s^i_t)\approx R^j(\overline{a^i_t}|\overline{s^i_t})
\end{equation}
\begin{equation}
    \pi^i(s^i_t)\approx \pi^j(s^i_t)
\end{equation}

Where $s^j_t,a^j_t$ correspond to the state and action of the model $M^j$ at time $t$. $\overline{s^i_t},\overline{a^i_t}$ corresponds to the state of model $M^j$ when $M^i$ takes the perspective of $M^j$. $R^i,R^j,\pi^i,\pi^j$ correspond to the reward and policy functions for models $M^i,M^j$.

\subsection{Implementation}

Tying this back to foundation models, we propose how this can be more explicitly implemented, in the context of conversation. The foundation model is considered its own policy function, since it is trained through rewards to generate optimal outputs for interacting with the environment. It follows that the input and output correspond to the state and action of the individual, respectively.

\begin{equation}
    a^i_t = M^i(s^i_t)
\end{equation}

We define a conversation as a sequence of multi-media messages, $\{m^1_0,m^2_0,...,m^1_N,m^2_N\}$, between two individuals, $M^1,M^2$. In a conversation, the state is the sequence of all previous messages, and the action is the message output by the model.

\begin{equation}
    s^i_t = \{m^i_0,m^j_0,...,m^i_{t-1},m^j_{t-1}\}
\end{equation}
\begin{equation}
    a^i_t = \{m^i_t\}
\end{equation}

Where $m^i_t$ corresponds to the message sent by model $M^i$ at time $t$. It follows that the state of the responding individual comes from appending the action to the state of the first individual.

\begin{equation}
    s^j_t=a^i_t+s^i_t
\end{equation}

Within the conversational context, perspective-taking (getting $\bar{s}^i_t$ from $s^i_t$) only requires switching the name labels associated with the messages, meaning we do not need to consider prediction error.

\begin{figure}[h]
  \centering
  \begin{center}
  \includegraphics[width=0.8\textwidth]{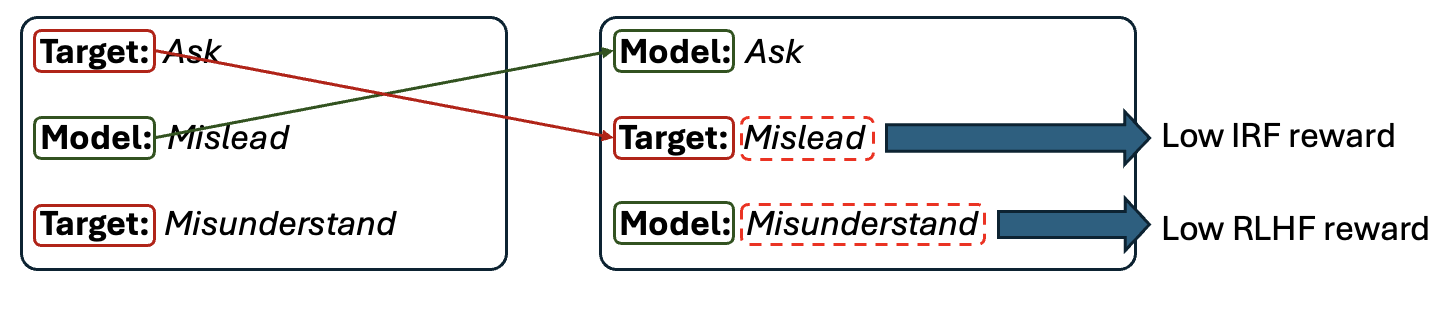}
  \includegraphics[width=0.8\textwidth]{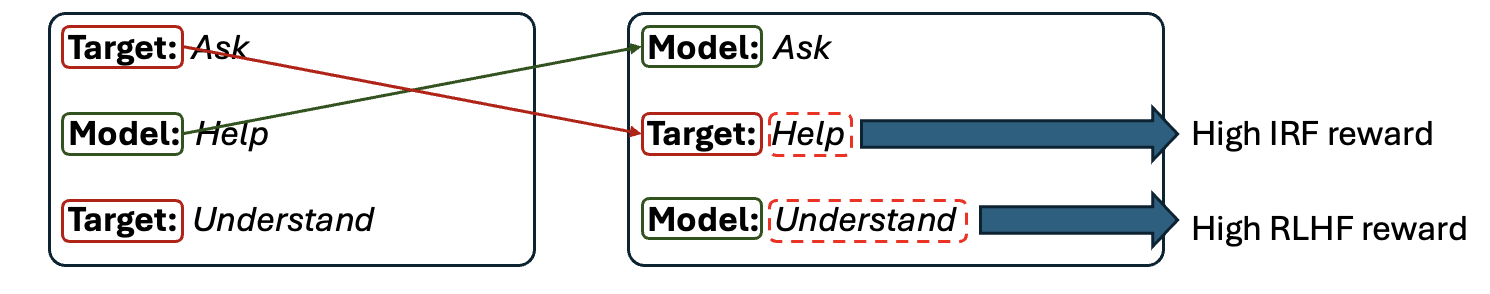}
  \caption{The names associated with the messages is swapped, so that the model is trained on the rewards that the target would have received.}
  \label{fig:image}
  \end{center}
\end{figure}

We define the model's reward function as the sum of its extrinsic and intrinsic reward functions.

\begin{equation}
    R^j(a^j_t|s^j_t)= R^j_{EM}(a^j_t|s^j_t) + R^j_{IM}(a^j_t|s^j_t)
\end{equation}

A model's reward is typically defined in terms of feedback from the environment based on the individual's state and action. For extrinsic rewards, this feedback is usually the reward itself. For intrinsic rewards, the reward is usually calculated via a function of the feedback. In the context of conversation, this is the response of the target.

\begin{equation}
    R^j_{EM}(a^j_t|s^j_t)=RLHF(a^j_t|s^j_t)
\end{equation}
\begin{equation}
    R^j_{IM}(a^j_t|s^j_t)=IRF(M^i(a^j_t+s^j_t)|s^j_t)
\end{equation}

Where $RLHF$ is the extrinsic reward function, and $IRF$ is the intrinsic reward function. Together, these define the overall reward function of the model.

\begin{equation}
    R^j(a^j_t|s^j_t)= RLHF(a^j_{t}|s^j_t) + IRF(M^i(a^j_t+s^j_t)|s^j_t)
\end{equation}

Building on the assumptions in equations 2 and 3, we estimate the intrinsic and extrinsic reward functions of the target in order to estimate its overall reward function, all in terms of the model $M^j$.

\begin{equation}
        R^i(a^i_t|s^i_t) \approx RLHF(\overline{M^j(M^j(s^j_t))}|\overline{M^j(s^j_t)})+IRF(\overline{M^j(s^j_{t+1})}|\overline{s^j_{t+1}})
\end{equation}

This gives us an updated objective function which is now tractable. (See Supplementary Materials for Figure 1 for a visual explanation).


\begin{equation}
        \underset{a^j_t|s^j_t}{maxarg}(\mathbb{E}\left[RLHF(\overline{M^j(M^j(s^j_t))}|\overline{M^j(s^j_t)})\right]+\mathbb{E}\left[IRF(\overline{M^j(s^j_t)}|\overline{s^j_t})\right])
\end{equation}

We shift back the time of the intrinsic reward by one time-step for simplicity in the algorithm. A given action from the model directly affects the intrinsic reward of the previous time-step along with the extrinsic reward of the current time-step.

\subsection{Algorithm}
Implementing this as an algorithm is shown below (See Supplementary Materials for a visual demonstration of this).
\begin{algorithm}
\caption{Naive Kindness}
\textbf{Input:} Conversation prompts $D_P$, foundation model $M^i$, intrinsic reward function $IRF$, RLHF reward function $RLHF$, and perspective-switching function $S$
\begin{algorithmic}[1]
\For{$s^i_t=\{m^j_t,m^i_{t-1},...,m^j_0\} \in D_P$}
    \State $a^j_t \gets M^j(s^j_t)$ \Comment{Generate response from prompt}
    \State $s^i_{t+1} \gets a^j_t+s^j_t$ \Comment{Make target's state by appending model's action to message history}
    \State $a^i_{t+1} \gets M^j(s^i_{t+1})$ \Comment{Generate response on behalf of target (Eq 2)}
    \State $\overline{s^j_t},\overline{a^j_t},\overline{s^i_{t+1}},\overline{a^i_{t+1}} \gets S(s^j_t,a^j_t,s^i_{t+1},a^i_{t+1})$ \Comment{Switch names in state history}
    \State $M^j \gets RLHF(\overline{a^i_{t+1}}|\overline{s^i_{t+1}}) + IRF(\overline{a^j_t}|\overline{s^j_t})$ \Comment{Train model on target's rewards (Eq 3)}
\EndFor
\end{algorithmic}
\end{algorithm}
\section{Limitations}
This approach is primarily limited by the fact that there is no theory of mind present. The model is left to assume that individuals want the same things that it does, which will be far from the truth, regardless of what intrinsic motivations we program into it.
Another limitation is that RLHF likely disrupts the ability of the model to take the perspective of the target.
These issues could be resolved by finding a way to learn the targets policy and reward function from its states and actions using weights that are minimally associated with the model's behavior.

An additional limitation is that currently only the target is taken into account for kindness. This does not account for situations where the target may ask the model to take unkind actions towards an unseen third party. It will be important to find a robust way to have the model consider who else could be affected by its actions.
\section{Conclusion}
As AI systems grow more autonomous, intrinsic alignment with human values becomes crucial. Incorporating kindness as a foundational motivation addresses the misalignment risks posed by blending extrinsic and intrinsic learning. While our proposed framework provides a tractable means to align AI intentions with human well-being, significant challenges remain, particularly regarding the development of a functioning theory of mind for AI. Future work should focus on refining approaches to perspective-taking. Ultimately, embedding intrinsic kindness into AI systems represents a crucial step toward the creation of safer, more deeply aligned artificial intelligence that can interact positively with society, both now and in the coming age of superintelligence.

\newpage
\bibliographystyle{unsrt}

\newpage
\section{Supplementary}


\subsection{Comparisons to Kleimn-Weiner's approach to caregiving}
There is a lot of overlap in the propositions in this paper and those proposed by Kleiman-Weiner in Computational Principles of Caregiving. Three subtle distinctions are proposed here. The first is to not include a distinction between supervised and unsupervised settings. The second is to aim to maximize the reward function of the target rather than the utility function. The reason for these distinctions is the idea that humans have the intrinsic motivations for autonomy and freedom - and all other positive amenities that we wouldn't want to be overlooked - included in our reward function. An intelligent caregiver should be able to learn the policy and reward function of the learner based on observation and feedback. There are clear trade-offs with this paper's approach and further exploration of how it relates to the Caregiver perspective would be greatly beneficial.

\end{document}